\def\year{2018}\relax
\documentclass[letterpaper]{article} 
\usepackage{aaai18}  
\usepackage{times}  
\usepackage{helvet}  
\usepackage{courier}  
\usepackage{url}  
\usepackage{graphicx}  

\usepackage[T1]{fontenc}    
\usepackage{url}            
\usepackage{booktabs}       
\usepackage{amsfonts}       
\usepackage{nicefrac}       
\usepackage{microtype}      
\usepackage{amsmath,bm}
\usepackage{multirow}
\usepackage{array}
\usepackage{booktabs}
\usepackage{rotating}
\usepackage{float}
\usepackage{wrapfig,lipsum,booktabs}
\DeclareMathOperator*{\argmin}{argmin}

\usepackage{times}
\usepackage{amsmath, amssymb, amsthm, caption,bm}
\numberwithin{equation}{section}

\usepackage{algorithm}
\usepackage{algpseudocode}
\usepackage{amsmath}
\usepackage{xcolor}

\begin{document}
%
\title{Zero-shot Learning via Shared-Reconstruction-Graph Pursuit}
\author{Bo Zhao$^{1}$, Xinwei Sun$^{2}$, Yuan Yao$^{3}$, Yizhou Wang$^{1}$\\
\small $^1$Nat'l Engineering Laboratory for Video Technology,\\
\small Key Laboratory of Machine Perception (MoE),\\
\small    Cooperative Medianet Innovation Center, Shanghai,\\
\small	 Sch'l of EECS, Peking University, Beijing, 100871, China\\
\small	$^2$School of Mathematical Science, Peking University, Beijing, 100871, China \\
\small	$^3$Hong Kong University of Science and Technology and Peking University, China\\
{\small  \{bozhao, 1301110047, Yizhou.Wang\} @pku.edu.cn, yuany@ust.hk}
}

\maketitle
\begin{abstract}
Zero-shot learning (ZSL) aims to recognize objects from novel unseen classes without any training data.
Recently, structure-transfer based methods are proposed to implement ZSL by transferring structural knowledge from the semantic embedding space to image feature space to classify testing images. However, we observe that such a knowledge transfer framework may suffer from the problem of the geometric inconsistency between the data in the training and testing spaces.
We call this problem as the \emph{space shift problem}. In this paper, we propose a novel graph based method to alleviate this space shift problem. Specifically, a Shared Reconstruction Graph (SRG) is pursued to capture the common structure of data in the two spaces. With the learned SRG, each  unseen class prototype (cluster center) in the image feature space can be synthesized by the linear combination of other class prototypes, so that testing instances can be classified based on the distance to these synthesized prototypes. The SRG bridges the image feature space and semantic embedding space.
By applying spectral clustering on the learned SRG, many meaningful clusters can be discovered, which interprets ZSL performance on the datasets.
Our method can be easily extended to the generalized zero-shot learning setting.
Experiments on three popular datasets show that our method outperforms other methods on all datasets. Even with a small number of training samples, our method can achieve the state-of-the-art performance.
\end{abstract}


\begin{figure*}
  \centering
  \includegraphics[width=0.65\textwidth]{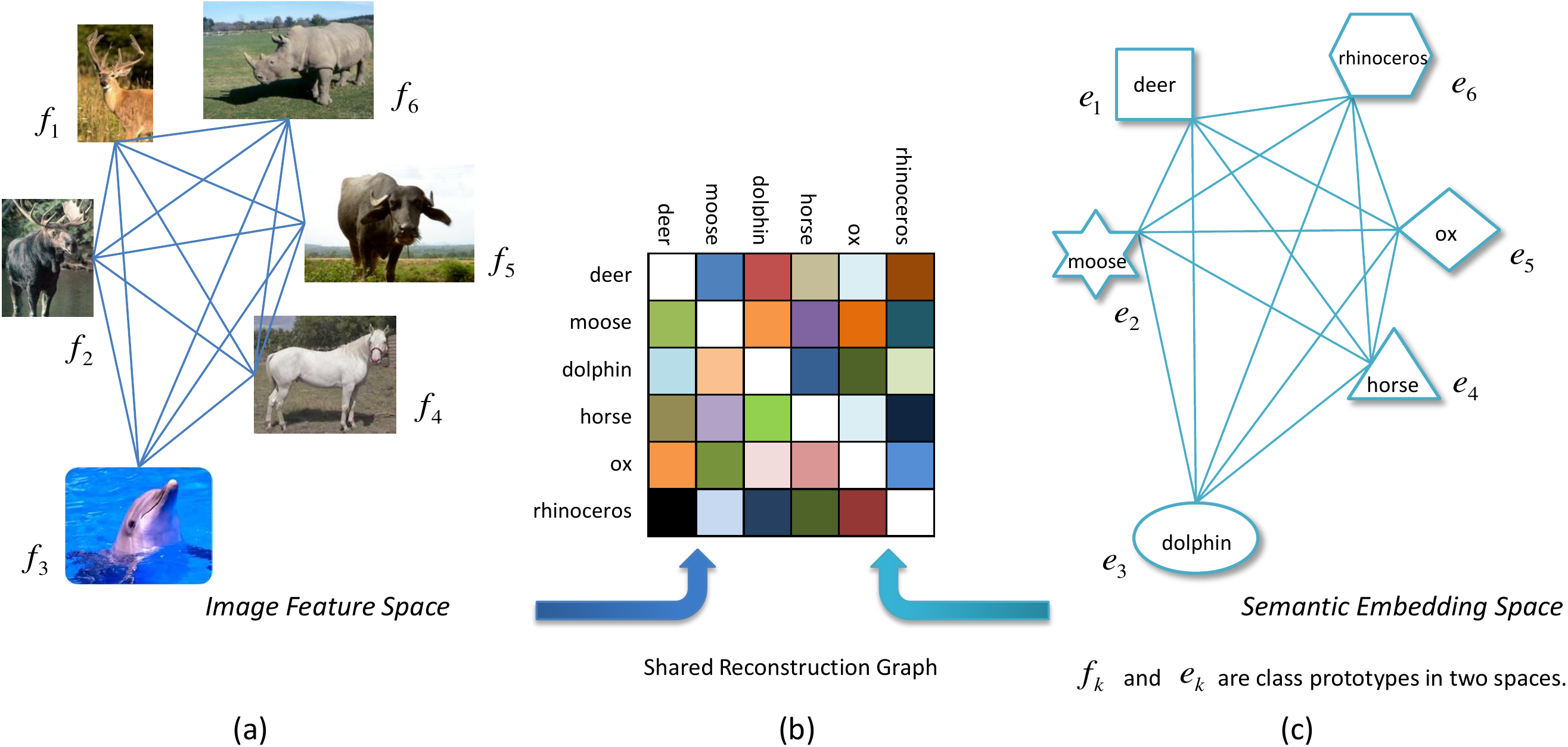}
  \caption{\small{Illustration of the proposed method. Each image prototype $f_k$ or semantic prototype $e_k$ (class-level data in the image feature and semantic embedding spaces) can be reconstructed by linear combination of other image prototypes or semantic prototypes. We aim to learn shared reconstruction coefficients between the two embedding spaces (also referred to as "spaces" for simplification), which means $f_k$ and $e_k$ of the same class share the same reconstruction coefficients. These coefficients form the Shared Reconstruction Graph (SRG). With the learned SRG, unseen image prototypes can be synthesized for classifying testing images.}}
  \label{fig:firstfig}
\end{figure*}

\section{Introduction}
In recent years, significant progress in Artificial Intelligence has been achieved by exploiting big training data and deep neural networks \cite{AlexNet}.
However, current image recognition tasks only focus on a small fraction of objects in the real world. For instance, ImageNet Large Scale Visual Recognition Challenge (ILSVRC) contains 1,000 popular categories in ImageNet for training and testing.
This recognition ability is far away from human beings' ability, as ordinary people can distinguish more than 30,000 basic level concepts \cite{biederman1987recognition}.
In addition, to learn a particular concept, popular deep networks require hundreds to thousands of labeled training data. It may be expensive even impossible to collect enough number of labeled training data for some categories, such as wild animals, rare plants and industrial products.
Hence, it is unrealistic to extend the recognition ability of machines only relying on collecting more training data. On the other hand, human beings have the ability to learn a new visual concept without seeing it. For example, a child can recognize the "giant panda" at the first glance, if he/she has learned from the description that giant pandas have black-and-white fur and eat bamboos. Inspired by such ability, zero-shot learning (ZSL) \cite{palatucci2009zero} aims to recognize instances from unseen classes by leveraging auxiliary knowledge. In this paper, we discuss zero-shot learning in the context of image recognition.

In zero-shot learning, images and corresponding labels of seen classes are provided for training. The trained model is then expected to recognize images from unseen classes. As training seen classes (source domain) and testing unseen classes (target domain) are disjointed, the auxiliary knowledge (e.g. attributes \cite{lampert2014attribute}, word vectors of labels \cite{socher2013zero}) is introduced to enable ZSL by knowledge transfer. Usually, images are embedded in the image feature space (using hand-crafted or deep feature extractors), and labels are embedded in the semantic (label) embedding space (using auxiliary knowledge, e.g. attributes or/and word vectors). In the semantic embedding space, unseen classes are semantically related to seen classes, so that knowledge transfer is possible. 

The most popular zero-shot learning framework \cite{lampert2014attribute} aims to learn the mapping function between the image feature space (F) and the semantic embedding space (E), i.e., F$ \rightarrow$E. An testing unseen image is first mapped to the semantic embedding space (using the learned mapping) then classified in this space using Nearest neighbor (NN) classifier  \cite{kodirov2017semantic} or Bayesian classifier \cite{lampert2014attribute}.
Recently, learning the mapping function in the reverse direction \cite{zhang2016learning}, i.e., E$\rightarrow$F, is advocated to relieve the hubness problem \cite{radovanovic2010hubs}.
In this mapping-transfer framework, the mapping function is learned on seen classes then transferred directly to unseen classes. They mainly suffer from "domain shift problem" \cite{fu2015transductive}, because the disjointed training and testing classes may have very different visual feature distribution.

Some works \cite{naha2015zero,changpinyo2016synthesized,wang2016relational,zhao2016zero} try to learn structural knowledge among semantic embeddings then transfer to the image features for synthesizing unseen image data or classification models. The motivation is that some structures among classes (such as manifold structure) are similar in the two spaces. Testing unseen images are classified based on these synthesized unseen data (using NN classifier) or classification models.
In such structure-transfer framework, the domain shift problem can be relieved by transferring structural knowledge from unseen semantic embeddings.

\begin{figure}
  \centering
  \includegraphics[width=0.47\textwidth]{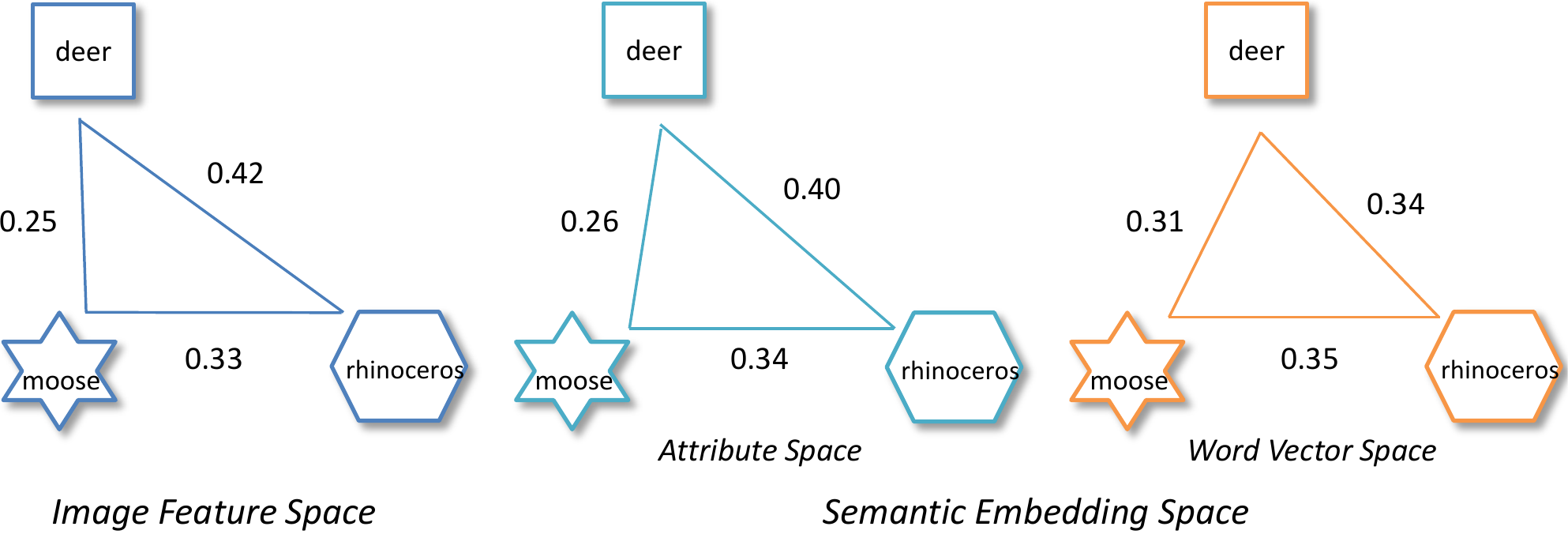}
  \caption{\small{Illustration of space shift problem. The weight on each edge is the normalized Euclidean distance between two embeddings.}}
  \label{fig:spaceshift}
\end{figure}

\textbf{Space Shift Problem} As these embedding spaces are built using inherently different data and methods, such structure-transfer framework mainly suffers from another kind of training / testing shift problem. We name it "\emph{space shift problem}" due to the shift of training and testing spaces. Specifically, the image feature space is built on image recognition task, the attribute space is built on human knowledge, while the word vector space is built on word embedding task.
The similarity between two classes differs in these embedding spaces. As shown in Fig. \ref{fig:spaceshift}, although "deer" and "moose" are closest in all three embedding spaces, the whole structures in these embedding spaces significantly differ.
Hence, the direct knowledge transfer from one embedding space to another will suffer from the problem caused by the shift of training and testing spaces, i.e., space shift problem.

In this paper, we also implement ZSL by transferring the structural knowledge in the two embedding spaces. For alleviating the space shift problem, we learn the shared structural knowledge which is adapted to both two spaces. Specifically, we aim to learn the Shared Reconstruction Graph (SRG) between the two spaces (illustrated in Fig. \ref{fig:firstfig}).
In this graph, each node is each class, and the edges (i.e. structural knowledge) are defined as the reconstruction coefficients. It is intuitive that each class shares some features (or attributes) with others, so that each class prototype (class-level datum) can be reconstructed by linearly regressing on others with corresponding reconstruction coefficients. SRG is compatible in two embedding spaces, i.e., the class prototype in the two spaces (image prototype and semantic prototype) can be reconstructed using the same reconstruction coefficients (weighted edges in SRG).
Finally, unseen class prototypes in the image feature space can be synthesized using SRG and testing images are classified based on distance to these synthesized prototypes.

We introduce the sparsity and locality regularization for selecting fewer, however, more related classes during the reconstruction process. So the learned graph is sparse.
Our method can alleviate the space shift problem, because we adapt SRG to both two spaces.
Many meaningful clusters on a dataset can be discovered by implementing clustering on the learned SRG, which makes our method more interpretable. These discovered clusters also help explain ZSL performance on different datasets.
Another benefit is that our method can achieve the state-of-the-art performance using a small number of training samples.
We also extend our method to general zero-shot learning, in which testing images may come from both seen and unseen classes.

The main contributions of this paper include: i) The space shift problem in zero-shot learning is defined. ii) The Shared Reconstruction Graph is proposed for implementing ZSL, which can alleviate the space shift problem. iii) An alternating optimization algorithm is proposed to learn SRG. iv) Many meaningful clusters on each dataset can be discovered via clustering on SRG, which are helpful to explain ZSL performance on the dataset.

\section{Related Work}
\subsection{Existing Problems in ZSL}
\subsubsection{Domain Shift Problem}
The mapping-transfer framework suffers from the domain shift problem \cite{fu2015transductive} due to the different visual feature distribution of seen and unseen classes. The domain shift problem blocks the effectiveness of some nonlinear models \cite{socher2013zero,romera2015embarrassingly}. As mentioned in \cite{zhang2016learning}, very few deep models exist in ZSL and the current state-of-the-art performances are achieved by some linear models.

For solving domain shift problem, \cite{fu2015transductive} propose the transductive setting for ZSL, i.e. unlabeled testing unseen images are provided for training. So transductive methods \cite{kodirov2015unsupervised,zhang2016SPZSL,deutschzero} can adapt to testing unseen images. However, the transductive setting sometimes is unrealistic and against the basic zero-shot learning setting.

\subsubsection{Space Shift Problem} \cite{changpinyo2016synthesized,wang2016relational,zhao2016zero,deutschzero} try to relieve the domain shift problem by leveraging the manifold structures in the two embedding spaces. They assume that the manifold structures in the two spaces are similar, so that the manifold of unseen semantic embeddings can be transferred for synthesizing virtual unseen image data \cite{wang2016relational} or models \cite{changpinyo2016synthesized}. As aforementioned, the direct structure transfer between the two spaces suffers from the space shift problem.
\cite{zhao2016zero} propose to adapt the synthesized unseen image data according to the distribution of unlabeled testing images in the transductive setting, while the transductive setting limits its generalization.


\subsection{Graph-based Methods for ZSL}
There exist some graph-based methods for ZSL, e.g., \cite{rohrbach2013transfer,fu2015transductive,fu2017zero}. Most of them only exploit the graph structure in one embedding space. \cite{rohrbach2013transfer} present a method based on label propagation on all testing instances in the image feature space. \cite{fu2015transductive} first embed all image features and semantic prototypes into a multi-view embedding space. Then labels of unseen instances are predicted by random walk on the hypergraph. \cite{fu2017zero}, however, focus on the label propagation on all seen semantic embeddings and formulate ZSL as an extended absorbing
Markov chain process. Different from them, our method exploit the graph structure among all classes in both two embedding spaces. We use the learned graph to synthesize unseen image data.

\subsection{Sparse Subspace Clustering}
We tend to explore the subspace structure rather than the manifold structure in the two spaces. Similar to Sparse Subspace Clustering (SSC) \cite{elhamifar2009sparse}, our method also contains a sparsity constraint. With this constraint, classes can be divided into many meaningful clusters which exist in different subspaces. We relax the affine constraint in SSC, so that linear subspaces are discovered. Motivated by Locality-constrained Linear Coding \cite{wang2010locality}, we introduce the locality constraint for large-scale datasets, because we want to leverage the similarity between classes to regularize the selection of data points.

\section{Methodology}
In zero-shot learning, images and labels of training seen classes are provided, i.e., $(\bm{X}^s,\bm{Y}^s)=\{(\bm{x}^s_1,{y}^s_1),...,(\bm{x}^s_{N^s},{y}^s_{N^s})\}$. For unseen classes, only the list of candidate labels $\bm{Y}^u$ are given. The label ${y}^u_i$ of each testing unseen image $\bm{x}^u_{i}$ is unknown.
Each datum $\bm{x}^s_{i}$ or $\bm{x}^u_{i}$ $\in \Re^{d \times 1}$ is a $d$-dimensional feature vector in the image feature space.
The label sets of the seen and unseen classes are disjointed, i.e. $ \bm{Y}^s \cap \bm{Y}^u = \emptyset$. Attributes or/and word vectors (auxiliary knowledge) are used as semantic (label) embeddings denoted as $\bm{E}^s=[\bm{e}^s_1,...,\bm{e}^s_{K^s}]$ and $\bm{E}^u=[\bm{e}^u_1,...,\bm{e}^u_{K^u}]$ for seen and unseen classes respectively. $\bm{e}^s_{k}$ and $\bm{e}^u_{k}$ $\in \Re^{p \times 1}$.
Using the seen data pairs $(\bm{x}^s_i,{y}^s_i)$, ZSL aims to predict the label ${y}^u_i$ for each testing unseen image $\bm{x}^u_i$ by leveraging the auxiliary knowledge $\bm{E}^s$ and $\bm{E}^u$ for knowledge transfer.

\subsection{Shared Reconstruction Graph}

\subsubsection{Class Prototype}
Usually, the class-level semantic embeddings (i.e. one attribute/word vector per class) are provided for implementing knowledge transfer. In our method, we focus on the class-level knowledge transfer.
Similar to existing works \cite{zhang2016SPZSL,wang2016relational}, we also assume that data from each class form a tight cluster and are separable from other classes in each embedding space.
Hence, each class can be represented by the cluster center, namely, the class prototype. Then the class prototype (class-level datum) of a seen class can be simply calculated by averaging all data from this class.
The class prototype in the image feature space, which is denoted as the image prototype $\bm{f}_{k}$ for simplification, can be calculated by ${\bm{f}}_k = \frac{1}{N_k}\sum{\bm{x}_i}, \; s.t. \; y_i = k, \ k \in \{1,...,K^{s}\}$.
The class prototype in the semantic embedding space (namely semantic prototype) has similar meaning. When only class-level semantic embeddings are provided, these semantic embeddings $\bm{e}_k$ serve as the semantic prototypes.
In this way, the class prototype in the image feature space and semantic embedding space form a datum pair $(\bm{f}_k, \bm{e}_k)$ for each class. Fig. \ref{fig:firstfig} (a) and (c) illustrate each image prototype and its corresponding semantic prototype in the two spaces.


\subsubsection{Graph Construction}
With the class-level data representation (class prototypes), it is natural to formulate the relationship among all classes using the graph structure. In the graph, each class prototype serves as the node, and the relationship between two classes is quantified as the weighted edge.

Different classes may share some features in the image feature space and semantic embedding space. For instance, in Animals with Attributes dataset, "giant panda" is black-and-white like "zebra", big like "polar bear", bulbous like "walrus". Hence, it is intuitive that a class like "giant panda" can be regressed using other classes. Inspired by this, we reconstruct each class prototype by linear combination of other ones with corresponding reconstruction coefficients.

In the semantic embedding space, we can reconstruct each semantic prototype using the following equation:
\begin{equation}
\label{equ:recon}
\bm{e}_k = \Sigma_{i=1}^K{\bm{e}_k^i\bm{\alpha}_k^i} = \bm{E}\bm{\alpha}_k, \; s.t.\; \bm{\alpha}_k^k = 0,
\end{equation}
where $\bm{\alpha}_k \in \Re^{K \times 1}$ is a column vector containing reconstruction coefficients. $K$ is the number of all classes. $\bm{\alpha}_k^i$ denotes the $i$th element in the vector $\bm{\alpha}_k$, and the constraint $\bm{\alpha}_k^k = 0$ means we use other ones to synthesize each semantic prototype. The coefficients $\bm{\alpha}_k$ measure the similarity between class $k$ and other base classes.

However, due to the lack of unseen image prototypes, the reconstruction coefficients for all classes in the image feature space cannot be learned directly. We aim to learn the shared reconstruction coefficients between the two spaces, so that the image prototype and semantic prototype of the same class can be synthesized using the same reconstruction coefficients.
These shared coefficients for all classes form the Shared Reconstruction Graph (SRG) which is denoted as $\bm{G}=(\bm{P},\bm{A})$. In this graph, each node in the node set $\bm{P}$ denotes each class prototype and all reconstruction coefficients compose the weight matrix $\bm{A} =[\bm{\alpha}_1, ..., \bm{\alpha}_K] \in \mathbb{R}^{K \times K}$. The graph is illustrated in Fig. \ref{fig:firstfig} (b).
Although unseen image prototypes are missing, the graph can be learned by sharing knowledge between the two spaces. The motivation is that some relationship among classes exists in both image feature and semantic embedding spaces, e.g. some visual attributes and semantic similarity.
With the learned SRG, we can synthesize these missing unseen image prototypes for classifying testing instances.

We expect the learned reconstruction coefficients are compatible in the image feature space, i.e., $\bm{f}_k = \bm{F}\bm{\alpha}_k,  \; s.t.\; \bm{\alpha}_k^k = 0$ with the same $\bm{\alpha}_k$ used in Equ. \ref{equ:recon}. We define the following reconstruction loss for learning SRG,
\begin{equation}
\begin{split}
L = \Sigma_k^K \| \bm{e}_k - \bm{E}\bm{\alpha}_k \|^2_F + \gamma \| \bm{f}_k - \bm{F}\bm{\alpha}_k \|^2_F,\\
s.t.\; \; \bm{\alpha}_k^k = 0, \; \gamma<1.    
\end{split}
\label{equ:loss}
\end{equation}
In this loss function, $\bm{F}= [\bm{F}^s, \bm{F}^u]$ contains both seen and unseen image prototypes. $\bm{F}^s=[\bm{f}^s_1,...,\bm{f}^s_{K^s}]$ and $\bm{F}^u=[\bm{f}^u_1,...,\bm{f}^u_{K^u}]$ are the sets of seen and unseen image prototypes respectively. $\bm{F}$ and $\bm{E}$ are placed in one-to-one correspondence. As unseen image prototypes are missing, we set the weight $\gamma$ for the reconstruction loss of image prototypes less than $1$, i.e. $\gamma<1$. We learn the two unknown variables $\bm{F}^u$ and $\bm{A}$ by minimizing the loss $L$, i.e.,
\begin{equation}
\label{equ:minimize}
\bm{F}^u, \bm{A} = \argmin_{\bm{F}^u, \bm{A}}{L}.
\end{equation}

\subsubsection{Comparison to Existing Methods}In our method, knowledge is transferred from semantic prototypes to image prototypes by sharing the reconstruction coefficients $\bm{A}$. Compared to existing mapping-transfer methods, unseen semantic embeddings are utilized during training process, so that the domain shift problem is relieved. Compared to existing structure-transfer methods, the graph is adapted to seen image prototypes, so that the space shift problem is alleviated.

\subsection{Sparsity and Locality Regularization}
Sparsity regularization is widely used in different optimization problems, e.g. Sparse Representation and Feature Selection.
It is obvious that only a little strong connection between classes exists in both image feature and semantic embedding spaces \cite{yoon2017combined,choi2016knowledge}. Other weak connection may only stand in one space. For example, in ImageNet dataset, "tiger" has a weak connection with "mud turtle", though the reconstruction coefficient can be calculated using Equ. \ref{equ:loss}. Such coefficient is meaningless even harmful to the prediction of unseen image prototypes.

We tend to use only a few classes with strong connection to synthesize some class. So we restrict the linear reconstruction by introducing sparsity regularization. For the sake of sparsity, we use L1-norm as the sparsity regularization, i.e., $\lambda\|\bm{D}_k\bm{\alpha}_k\|_1$, where $\lambda$ is the sparsity parameter and $\bm{D}_k$ is a diagonal matrix. Therefore, the regularized loss function is
\begin{equation}
\begin{split}
L = \Sigma_k^K \| \bm{e}_k - \bm{E}\bm{\alpha}_k \|^2_F + \gamma \| \bm{f}_k - \bm{F}\bm{\alpha}_k \|^2_F \\
+ \lambda\|\bm{D}_k\bm{\alpha}_k\|_1,
\;  \; \;  \; s.t.\;  \; \bm{\alpha}_k^k = 0.     
\end{split}
\label{equ:loss_full}
\end{equation}

For small-scale datasets, each $\bm{D}_k$ for $k$th class is set to be an identity matrix. With this sparsity regularization, some weakly related classes can be removed and the corresponding reconstruction coefficients will be zero.

For large-scale datasets (e.g. ImageNet), some classes are too far away to contribute to the reconstruction of a particular class. For instance, "tiger" and "umbrella" has little relationship in any aspects.
It is intuitive that human beings prefer to find relationship between close concepts. Hence, we further introduce the locality regularization for large-scale datasets. Close classes rather than distant ones are expected to be selected.
We introduce locality regularization by setting the diagonal matrix $\bm{D}_k$. Each diagonal element $\bm{D}_k^{i,i}$ (namely, locality penalty) is set based on the distance (or dissimilarity) between $k$th and $i$th classes,
\begin{equation}
\begin{split}
\bm{D}_k^{i,i} =
\begin{cases} g ( \bm{e}_i, \bm{e}_k ) \, \; \; \; if \;\; i \neq k, \\
1, \; \; \; \; else.
\end{cases}
\end{split}
\label{equ:localpen}
\end{equation}
The function $g ( \cdot, \cdot )$ is an increasing function of the distance (or dissimilarity), and $g ( \cdot, \cdot ) > 0$. $\bm{D}_k^{k,k}$ is set to be 1 for avoiding calculation problems.

The reconstruction coefficients $\bm{A}$ and unseen image prototypes $\bm{F}^u$ are learned simultaneously by minimizing this regularized loss (Equ. \ref{equ:loss_full}).

\subsection{Optimization}
Although the loss function (Equ. \ref{equ:loss_full}) is not convex for $\bm{F}^u$ and $\bm{A}$ simultaneously, it is convex for each variable respectively. Hence, we present an alternating optimization algorithm for solving it.
First, we fix $\bm{F}^u$ and optimize $\bm{A}$,
\begin{equation}
\label{equ:opt_alpha}
\begin{split}
\min_{\bm{A}} L = \min_{\bm{A}} \Sigma_k^K \|
\begin{array}{c}
\bm{e}_k - \bm{E}\bm{\alpha}_k\\
\gamma(\bm{f}_k - \bm{F}\bm{\alpha}_k)
\end{array}
\|^2_F \\
+ \lambda\|\bm{D}_k\bm{\alpha}_k\|_1,
\; \; \; \;  s.t.\; \; \bm{\alpha}_k^k = 0.       
\end{split}
\end{equation}
We denote $\bm{\beta}_k = \bm{D}_k\bm{\alpha}_k$, so $\bm{\alpha}_k = (\bm{D}_k)^{-1}\bm{\beta}_k$. Then Equ. \ref{equ:opt_alpha} can be written as
\begin{equation}
\label{equ:opt_alpha_final}
\begin{split}
\min_{\bm{A}} L = \min_{\bm{A}} \Sigma_k^K \|
[
\begin{array}{c}
\bm{e}_k\\
\gamma\bm{f}_k
\end{array}
] -
[
\begin{array}{c}
\bm{E}(\bm{D}_k)^{-1}\\
\gamma\bm{F}(\bm{D}_k)^{-1}
\end{array}
]
\bm{\beta}_k
\|^2_F \\ + \lambda\|\bm{\beta}_k\|_1,
\; \; \; \; s.t.\; \;  \bm{\beta}_k^k = 0.
\end{split}
\end{equation}

The Equ. \ref{equ:opt_alpha_final} is a typical LASSO problem, and $\bm{\beta}_k$ can be easily solved by many available solvers, e.g., LeastR \cite{liu2009efficient}. So $\bm{\alpha}_k$ is solved.
As the $\bm{F}^u$ is unknown in the first iteration, we set $\gamma = 0$ during the first iteration.

Second, we fix $\bm{A}$ and optimize $\bm{F}^u$. When $\bm{A}$ is fixed, it is equal to optimize the following equation:
\begin{equation}
\label{equ:opt_Fu}
\begin{split}
\min_{\bm{F}^u} L = \min_{\bm{F}^u} \Sigma_k^K \gamma \| \bm{f}_k - \bm{F}\bm{\alpha}_k \|^2_F \\
= \min_{\bm{F}^u} \gamma \| \bm{F}(\bm{I}-\bm{A}) \|^2_F,
\end{split}
\end{equation}
where $\bm{I}$ is the identity matrix with the same size as $\bm{A}$.
With the denotation $\bm{\theta} = (\bm{I} - \bm{A})$, the Equ. \ref{equ:opt_Fu} is simplified:
\begin{equation}
\label{equ:opt_Fu_sec}
\begin{split}
\min_{\bm{F}^u} \gamma \| \bm{F}\bm{\theta} \|^2_F
= \min_{\bm{F}^u} \gamma \| [\bm{F}^s, \bm{F}^u][
\begin{array}{c}
\bm{\theta}^s\\
\bm{\theta}^u
\end{array}
]\|^2_F \\
= \min_{\bm{F}^u} \gamma \| \bm{F}^s \bm{\theta}^s + \bm{F}^u \bm{\theta}^u \|^2_F.
\end{split}
\end{equation}
$\bm{\theta}^s$ and $\bm{\theta}^u$ means that we divide $\bm{\theta}$ into two parts with corresponding sizes. Therefore, when
\begin{equation}
\label{equ:opt_Fu_final}
\bm{F}^u = - \bm{F}^s \bm{\theta}^s(\bm{\theta}^u)^{-1},
\end{equation}
the loss function obtains its minimum.
We repeat these two steps until both of them converge. This alternating optimization algorithm is shown in Alg. \ref{Algo:Opt}.

\begin{algorithm}[h]
  \caption{Learning Shared Reconstruction Graph}
  \begin{algorithmic}[1]
  \Require
      Seen image prototypes $\bm{F}^s$, all semantic embeddings $\bm{E}^s$ and $\bm{E}^u$;
  \Ensure
      Synthesized unseen image prototypes $\bm{F}^u$ and all reconstruction coefficients $\bm{A}$;
    \State $\textbf{Initialize:}$ Set $\gamma$ and $\delta$;
    \State Construct $\bm{D}_k$ for each class;
    \While{not converge}
    	 \State Update $\bm{A}$ by solving Equ. \ref{equ:opt_alpha_final},
    	 \State Update $\bm{F}^u$ using Equ. \ref{equ:opt_Fu_final},
    \EndWhile
  \end{algorithmic}
  \label{Algo:Opt}
\end{algorithm}

\subsection{Zero-shot Classification}
In zero-shot classification, testing instances are from unseen classes. In our method, testing unseen instances can be classified based on distance to these synthesized unseen image prototypes $\bm{F}^u$. Similar to many existing works \cite{palatucci2009zero,kodirov2017semantic}, we choose the simple Nearest Neighbor classifier to classify testing instances. The label of each testing unseen instance is predicted as the one with the minimum distance, i.e.
\begin{equation}
{y}_i^u = \argmin_k{\|\bm{x}_i^u-{\bm{f}}_k^u\|_F}, 
\end{equation}
where ${\bm{f}}_k^u$ means each synthesized unseen image prototype. 

Our method can be easily extended to a more general setting of generalized zero-shot learning (GZSL) \cite{chao2016empirical}. GZSL means that a testing instance may come from any seen or unseen class, i.e., ${y}_i \in \bm{Y}^s\cup\bm{Y}^u$. The only modification is that, for GZSL, we measure the distance to both seen and unseen classes, i.e., ${y}_i = \argmin_{k}{\|\bm{x}_i-{\bm{f}}_k\|_F}$.

\subsection{Subspace Clustering}
When $D_{k}$ is an identity matrix in Equ. \ref{equ:loss_full}, the loss function is similar to the sparse representation function in Sparse Subspace Clustering (SSC) algorithm \cite{elhamifar2009sparse}, which assumes each datum is drawn from a linear subspace $\mathcal{S}$ with a basis $\mathcal{U}$. In our problem, the basis $\mathcal{U}$ means the class prototype.
Different from SSC, the affine combination constraint is relaxed in our method.
By constructing the balanced graph $\bm{\tilde{G}} = (\bm{V}, \bm{\tilde{A}})$, with $ \bm{\tilde{A}} = \bm{A} + \bm{{A}}^T$, we can implement spectral clustering on all classes. We calculate the Laplacian matrix $L_p$ of the balanced graph $\bm{\tilde{G}}$ and apply K-means clustering to the $n$ eigenvectors of $L_p$ corresponding to the smallest $n$ eigenvalues. Then, all classes in a dataset are divided into many clusters.

It's expected that classes share more similar features should be divided into the same subspace, which indicates the good interpretability of our method. In experiments, we illustrate how our method discovers many meaningful latent clusters in a dataset.

\section{Experiments}
\subsection{Datasets \& Settings}
\subsubsection{Datasets} We evaluate our method on three popular datasets, namely Animals with Attributes (AwA) \cite{lampert2009learning}, Caltech-UCSD Birds-200-2011 (CUB) \cite{WahCUB_200_2011} and ImageNet \cite{deng2009imagenet}. AwA is a coarse-grained dataset which contains images of 50 kinds of common animals. 10 classes are selected as the unseen classes, and the rest are seen classes. 85-dim attributes are provided. CUB is a fine-grained dataset that contains 200 kinds of birds. 50 classes are used as the unseen classes. The rest 150 classes are seen classes. 312-dim attributes are provided. We follow the seen/unseen splits of AwA and CUB used in \cite{wang2016relational}.
ImageNet 2012/2010 is a large-scale dataset. No attributes are provided in this dataset. Following \cite{fu2016semi}, we use 1,000 classes in ImageNet 2012 as seen classes. 360 classes in ImageNet 2010 which do not exist in ImageNet 2012 serve as unseen classes.
We compare to state-of-the-art methods under the inductive setting, because the transductive setting is sometimes unrealistic.
%

\begin{table*}[]
\centering
\small
\begin{tabular}{c|ccc|ccc|cc}
\toprule
                  & \multicolumn{3}{c|}{AwA} & \multicolumn{3}{c|}{CUB} & \multicolumn{2}{c}{ImageNet} \\ \hline
S/L				  & A               & W               & AW              & A              & W                & AW             & W - Top 1         & W - Top 5         \\ \hline
1/1               & -               & -               & -               & -              & -                & -              & \textbf{8.14}     & \textbf{18.26}     \\ \hline
1/0               & \textbf{81.22}  & \textbf{73.19}  & \textbf{83.62}  & \textbf{54.31} & \textbf{49.03}   & \textbf{58.10} &  7.72             & 17.79    \\ \hline
0/0               & 64.93           & 64.43           & 81.22           & 46.55          & 47.33            & 56.76          &  3.38             & 9.23             \\ \bottomrule
\end{tabular}
\caption{\small{Analysis of sparsity and locality regularization. S/L denotes sparsity/locality. 1 means with this regularization term while 0 means without this term. Clearly, sparsity and locality regularization terms are both important for improving ZSL performance (\%).} The result of AwA and CUB with locality regularization is omitted, because locality regularization is only for large-scale datasets.}
\label{tab:analysis_SL}
\end{table*}

\begin{table*}[]
\centering
\small
\begin{tabular}{l|cccc|cccc}
\toprule
\multirow{2}{*}{} & \multicolumn{4}{c}{AwA}                                                                   & \multicolumn{4}{|c}{CUB}                                                                   \\ \cline{2-9}
                  & \multicolumn{1}{c}{${u\rightarrow u}$} & \multicolumn{1}{c}{${s\rightarrow s}$} & \multicolumn{1}{c}{${u\rightarrow \tau}$} & \multicolumn{1}{c}{${s\rightarrow \tau}$} & \multicolumn{1}{|c}{${u\rightarrow u}$} & \multicolumn{1}{c}{${s\rightarrow s}$} & \multicolumn{1}{c}{${u\rightarrow \tau}$} & \multicolumn{1}{c}{${s\rightarrow \tau}$} \\ \hline
DAP/IAP           & 51.1 / 56.3          & 78.5 / 77.3          & 2.4 / 1.7            & 77.9 / 76.8          & 38.8 / 36.5          & 56.0 / 69.6          & 4.0 / 1.0            & 55.1 / 69.4          \\ \hline
ConSE             & 63.7                 & 76.9                 & 9.5                  & 75.9                 & 35.8                 & 70.5                 & 1.8                  & 69.9                 \\ \hline
SC\_struct        & 73.4                 & 81.0                 & 0.4                  & 81.0                 & 54.4                 & \textbf{73.0}        & 13.2                 & \textbf{72.0}        \\ \hline
Ours              & \textbf{83.62}       & \textbf{85.92}       & \textbf{38.46}       & \textbf{83.23}       & \textbf{58.10}       & 69.56                & \textbf{24.62}       & 66.53                \\ \bottomrule
\end{tabular}
\caption{\small{Assessing GZSL methods on the four performance metrics (\%).}}
\label{tab:GZSL}
\end{table*}

\subsubsection{Image \& Semantic Embedding} For coarse-grained datasets (AwA and ImageNet), we use image features extracted by VGG-19 \cite{simonyan2014very}. Attributes and 500-dim word vectors are used as semantic embeddings. For fine-grained dataset (CUB), we use GoogLeNet \cite{szegedy2015going} + ResNet \cite{ResidualNet} features. Attributes and 1024-dim word vectors are used as semantic embeddings.

\subsubsection{Parameters Selection}
There are only two free parameters, namely $\lambda$ and $\gamma$, in the loss function. We select these parameters by Cross-Validation. Specifically, we split the seen classes into 5 folds for keeping the same seen/unseen ratio. Then one fold is used as new "unseen" classes, and the rest are "seen" classes. The parameters are selected based on the average performance on each fold. The searching range of $\lambda$ and $\gamma$ are $10^{[-2:2]}$ and $10^{[-2:0)}$ respectively. For large-scare dataset (ImageNet), we choose $g ( \bm{e}_i, \bm{e}_k ) = \log{(1+\|\bm{e}_i, \bm{e}_k\|_F)} $ to calculate the penalty value.

\subsection{Comparison to State-of-the-Art}
\subsubsection{Small-scale Datasets}
There are many works which are evaluated on AwA and CUB datasets.
We compare to some state-of-the-art ZSL methods, namely, DAP/IAP \cite{lampert2014attribute}, SJE \cite{akata2015evaluation}, SC\_struct \cite{changpinyo2016synthesized}, LatEm \cite{xian2016latent}, LEESD \cite{dinglow}, SS-Voc \cite{fu2016semi}, DCL\cite{guo2017zero} and JLSE \cite{zhang2016zero}. We also repeat experiments of ESZSL \cite{romera2015embarrassingly} and RKT \cite{wang2016relational} using the same data (image features and semantic embeddings).

It is clear from Tab. \ref{tab:comparison} that our method achieves the best performances (namely \textbf{83.62\%} and \textbf{58.10\%}) on both AwA and CUB. Our method outperforms the runner-up methods (RKT on AwA and LEESD on CUB) by 2.21\% and 1.90\% on the two datasets respectively. The improvement, compared to RKT, proves the effectiveness of the adaptation on both image feature and semantic embedding spaces. In other words, our method can alleviate the space shift problem.

\begin{table}[t]
\centering
\small
\begin{tabular}{l|c|c|c|c}
\toprule
    \multirow{2}{*}{Methods} &\multirow{2}{*}{AwA}  &\multirow{2}{*}{CUB}  &\multicolumn{2}{c}{ImageNet } \\
    &                 &              &     Top-1      & Top-5\\ \hline
    DAP/IAP           	& 41.4/42.2$^\dagger$       & -            &     &          \\ 
    SJE              	& 66.7          & 50.1         &     &          \\ 
    SC\_struct          & 72.9          & 54.7         &     &     \\ 
    LatEm               & 76.1            & 47.4         &     &     \\ 
    LEESD			 	& 76.6          & 56.2         &     &     \\ 
    SS-Voc            	& 78.3$^\ddagger$  & -          & \textbf{9.5}$^\ddagger$     & 16.8$^\ddagger$     \\ 
    ConSE          		& -              & -            & 7.8                      & 15.5    \\ 
    DeViSE          	& -               & -            & 5.2                 & 12.8     \\ 
    DCL					& 79.07          & -      &      &      \\ 
    JLSE              	& 80.46           & 42.11        &     &     \\ \midrule
    \emph{ESZSL}        & 79.53           & 51.90        &     &     \\ 
    \emph{RKT}          & 81.41           & 55.59        &     &     \\
    Ours (SRG)          & \textbf{83.62}   & \textbf{58.10}    & {8.14}  & \textbf{18.26}  \\   \bottomrule
\end{tabular}
\caption{\small{Comparison to the state-of-the-art (\%). $^\dagger$ denotes that hand-crafted features are used. $^\ddagger$ means that extra vocabulary knowledge (nearly 310k word vectors) is utilized, which is not fair to compare with. \emph{ESZSL} and \emph{RKT} mean re-implemented experimental results.}}
\label{tab:comparison}
\end{table}

\subsubsection{Training with Small Samples}
\begin{figure}
  \centering
  \includegraphics[width=0.4\textwidth]{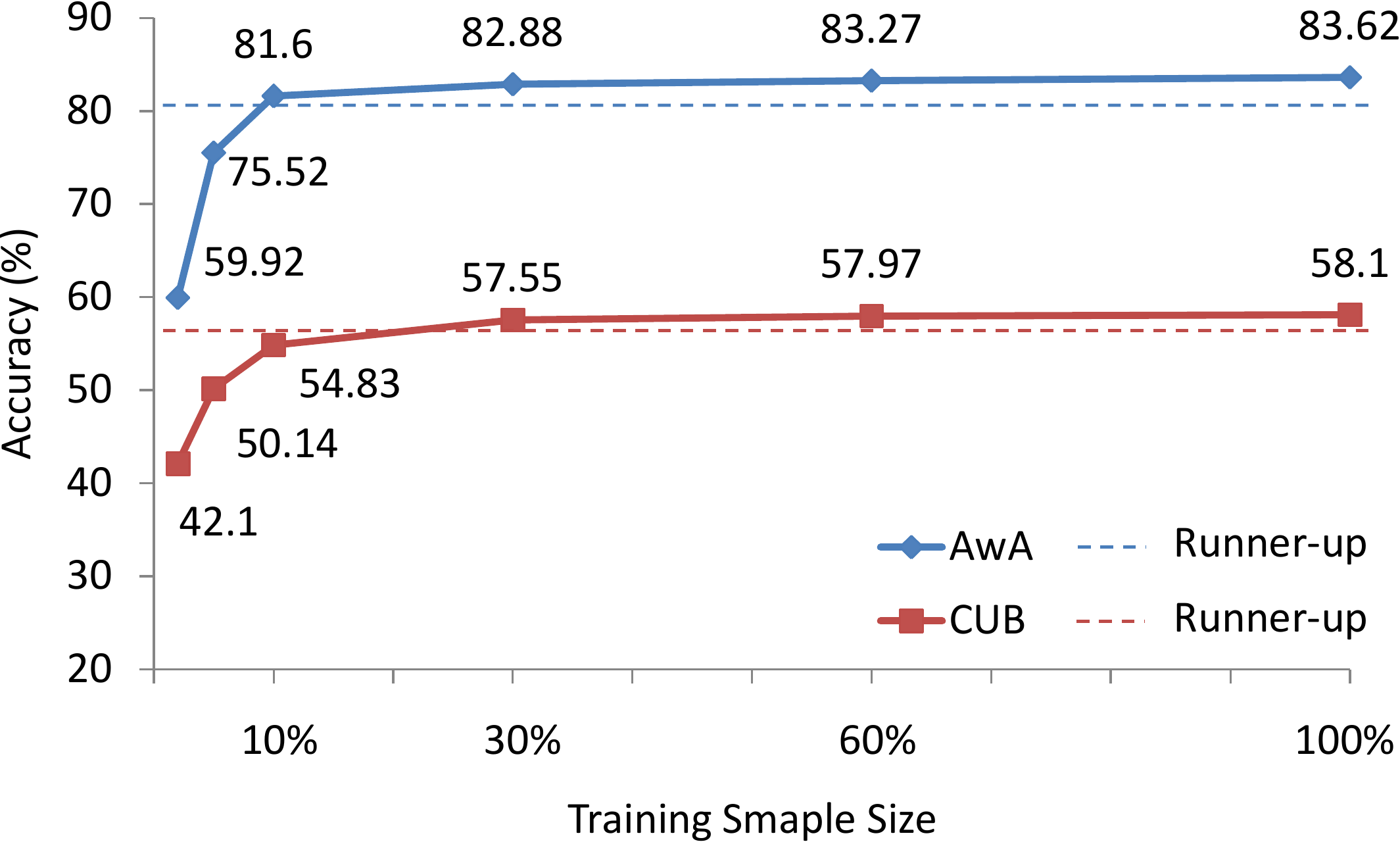}
  \caption{\small{Classification accuracies using different sizes of training samples. Specifically, 1, 3 images per class, 10\%, 30\%, 60\% and 100\% training samples are used for training. The blue and red dashes denote the runner-up performance on AwA and CUB respectively.}}
  \label{fig:sample}
\end{figure}

One benefit of our method is that a reliable Shared Reconstruction Graph for classification can be learned with a small number of training samples. We measure the classification accuracies of our method with different sizes of training seen images, namely, 1 image per class, 3 images per class, 10\%, 30\%, 60\% and 100\% of original samples. All samples are randomly selected. As shown in Fig. \ref{fig:sample}, our method can outperform the runner-up (RKT 81.41\%) on AwA using only 10\% samples and the runner-up (LEESD 56.2\%) on CUB using only 30\% samples. It should be noticed that the average sample size on CUB is 60 per class and 30\% samples means around 18 images.

\begin{table*}[]
\centering
\small
\begin{tabular}{llllllll}
\toprule
1  & beaver        & walrus     & otter           & \textbf{seal}           &             &             &             \\ \hline
2  & killer whale  & blue whale & dolphin         & \textbf{humpback whale} &             &             &             \\ \hline
3  & antelope      & moose      & deer            & giraffe        & zebra       & horse       &             \\ \hline
4  & lion          & bobcat     & tiger           & \textbf{leopard}        &             &             &             \\ \hline
5  & mouse         & hamster    & squirrel        & mole           & rabbit      & sheep       &             \\ \hline
6  & elephant      & rhinoceros & \textbf{hippopotamus}    &                &             &             &             \\ \hline
7  & buffalo       & cow        & ox              & \textbf{pig}            &             &             &             \\ \hline
8  & skunk         & \textbf{raccoon}    & \textbf{rat}      &       &             &             &             \\ \hline
9  & collie        & dalmatian  & German shepherd & chihuahua  & Siamese cat & \textbf{Persian cat} & \textbf{giant panda} \\ \hline
10 & fox           & weasel     & wolf            & grizzly bear   & polar bear  & bat         &             \\ \hline
11 & spider monkey & gorilla    & \textbf{chimpanzee}&                &             &             &             \\ \bottomrule
\end{tabular}
\caption{\small{Clustering result on AwA. Bold ones are unseen classes. It is clear that these clusters are meaningful and separable from each other. }}
\label{tab:clusters}
\end{table*}

\subsubsection{Large-scale Dataset}
The large number of categories makes zero-shot classification more difficult on ImageNet dataset. Only a few ZSL methods are evaluated on ImageNet dataset. We compare to state-of-the-art methods, namely, SS-Voc \cite{fu2016semi}, ConSE \cite{norouzi2013zero}, DeViSE \cite{frome2013devise}. The result is shown in Tab. \ref{tab:comparison}.
Although extra vocabulary knowledge is utilized in SS-Voc, our method still achieve the best performance (\textbf{18.26\%}) on ImageNet measured on Top-5 classification accuracy. When measured on Top-1 classification accuracy, our method obtain 8.14\% classification accuracy, which is better than ConSE and DeViSE.

\subsection{Analysis of Sparsity and Locality}
In this subsection, we analyze the effectiveness of sparsity and locality regularization in the objective function (Equ. \ref{equ:loss_full}). We calculate the classification accuracy on three datasets with different regularization terms, namely "sparsity + locality", "only sparsity" and "no regularization". For "only sparsity" experiment, each $\bm{D}_k$ is defined as an identity matrix. For "no regularization" experiment, we set $\lambda$ to be zero, so that $\bm{D}_k$ is disabled. We apply the locality regularization only on the large-scale dataset, i.e., ImageNet.

As shown in Tab. \ref{tab:analysis_SL}, the classification performance has a stable improvement on all three datasets when the sparsity regularization is introduced. When only single type of semantic embeddings is used, namely no fusion between attributes and word vectors, the improvement brought by the sparsity regularization is remarkable. On AwA, the improvement is 16.29\% and 8.76\% using attributes and word vectors respectively. On CUB, the improvement is 7.76\% using attributes as the semantic embeddings. The Top-1 and Top-5 classification accuracies on ImageNet both increase significantly due to the introduce of sparsity regularization.

When the locality regularization is introduced to ImageNet dataset, the classification accuracies on Top-1 and Top-5 measurement rise from 7.72\% to 8.14\% and 17.79\% to 18.26\% respectively. This improvement proves the effectiveness of the locality regularization for large-scale dataset.

\subsection{Generalized Zero-shot Learning}
The generalized zero-shot learning considers a more realistic setting that testing instances belong to both seen and unseen classes. In this GZSL experiment, we follow the setting presented in \cite{chao2016empirical} and assess our method on the following performance metrics: ${u\rightarrow u}$ the accuracy of classifying testing data from $\bm{Y}^u$ into $\bm{Y}^u$, ${s\rightarrow s}$ the accuracy of classifying testing data from $\bm{Y}^s$ into $\bm{Y}^s$, and ${u\rightarrow \tau}$ and ${s\rightarrow \tau}$ the accuracies of classifying testing data from $\bm{Y}^u$ or $\bm{Y}^s$ into the whole label list $\bm{Y}^u\cup\bm{Y}^s$. The two settings (${u\rightarrow u}$ and ${s\rightarrow s}$) are also viewed as the conventional ZSL and multi-class classification tasks.

We compare to the reported results in \cite{chao2016empirical} of three state-of-the-art methods, namely, DAP/IAP \cite{lampert2014attribute}, ConSE \cite{norouzi2013zero} and SC\_struct \cite{changpinyo2016synthesized}. It is clear from Tab. \ref{tab:GZSL} that our method achieves the best classification accuracies on all performance metrics on AwA. On CUB dataset, SC\_struct outperforms our method only when testing instances are from seen classes, i.e., ${s\rightarrow s}$ and ${s\rightarrow \tau}$. It indicates that our method gives priority to unseen classes. Overall, our method achieves state-of-the-art performance in the GZSL setting.

\subsection{Discovering Latent Clusters}
We illustrate the clustering result on AwA dataset. The reason is that CUB and ImageNet datasets have too many classes to display. In this experiment, attributes serve as the semantic embeddings.
We can set different cluster numbers in K-means algorithm and discover different kinds of clustering results. Tab. \ref{tab:clusters} shows the clustering result when the cluster number equals to 11.
It is clear that these clusters are meaningful and separable from each other. For example, animals in Cluster\_1 are aquatic animals except whales, while Cluster\_2 are the whales (dolphin belongs to whales). Cluster\_4 contains 4 fierce Felidae species. Cluster\_11 includes 3 kinds of primates. Other clusters also have clear meanings. This clustering result also verifies the good ZSL performance on AwA, because unseen classes (bold ones in Tab. \ref{tab:clusters}) all have close seen classes in the same cluster, which enable effective knowledge transfer.

\section{Conclusion}
In this paper, we define the space shift problem in ZSL. The Shared Reconstruction Graph is proposed for alleviating the space shift problem in ZSL. Unseen image prototypes are synthesized for classifying testing instances using the learned SRG. We introduce the sparsity and locality regularization for selecting fewer, however, more related classes in the reconstruction process. We also extend our method to GZSL. Many meaningful latent clusters on AwA are discovered by implementing clustering on SRG. Experiments show that our method outperforms the state-of-the-art even with a small number of training samples.

\bibliographystyle{named}
\bibliography{AAAI}

\end{document}